# EEG machine learning with Higuchi's fractal dimension and Sample Entropy as features for successful detection of depression

Milena Čukić[1,2*], David Pokrajac[3], Miodrag Stokić[4,5], Slobodan Simić[6], Vlada Radivojević[6], and Miloš Ljubisavljević[7]

[1] General Physiology with Biophysics, University of Belgrade, Belgrade, Serbia

[2] Amsterdam Health and Technology Institute, HealthInc, Amsterdam, the Netherlands

[3] Office of Institutional Effectiveness, Delaware State University, USA

[4] Life Activities Advancement Center, Belgrade, Serbia

[5] Institute for Experimental Phonetics and Speech Pathology, Belgrade, Serbia

[6] Institute for Mental Health, Belgrade, Serbia

[7] Department of Neurophysiology, College of Medicine and Health Sciences, UAE University, Al Ain, UAE

[*]Correspondence concerning this article should be addressed to Milena Čukić, Ph.D., e-mail: milena.cukic@gmail.com

**ABSTRACT**

Reliable diagnosis of depressive disorder is essential for both optimal treatment and prevention of fatal outcomes. In this study, we aimed to elucidate the effectiveness of two non-linear measures, Higuchi's Fractal Dimension (HFD) and Sample Entropy (SampEn), in detecting depressive disorders when applied on EEG. HFD and SampEn of EEG signals were used as features for seven machine learning algorithms including Multilayer Perceptron, Logistic Regression, Support Vector Machines with the linear and polynomial kernel, Decision Tree, Random Forest, and Naïve Bayes classifier, discriminating EEG between healthy control subjects and patients diagnosed with depression. We confirmed earlier observations that both non-linear measures can discriminate EEG signals of patients from healthy control subjects. The results suggest that good classification is possible even with a small number of principal components. Average accuracy among classifiers ranged from 90.24% to 97.56%. Among the two measures, SampEn had better performance. Using HFD and SampEn and a variety of machine learning techniques we can accurately discriminate patients diagnosed with depression vs controls which can serve as a highly sensitive, clinically relevant marker for the diagnosis of depressive disorders.

**INTRODUCTION**

Depression is forecasted to be the second cause of disability and death before 2030 world wide (Mathews and Loncar, 2006; Murray et al., 2010; World Organization of Mental health, 2012; World Health Organization, 2017). The effects of depression tend to extend beyond the individual patient, negatively impacting patients' immediate social environment. In the Netherlands alone, depression was the second most prevalent cause of a sick leave in 2016, making up some 11% of workforce loses (WHO Europe, Data and resources, 2014). The underlying pathogenesis of depression is still not known. Nevertheless, considerable evidence from neuroimaging studies shows structural and functional changes, which affect different brain regions and neurotransmitter systems (Arnone et al., 2012; Arnone et al., 2013; Koolschijn et al., 2009; Kwaasteniet et al., 2013; Vederine et al., 2011).

In present clinical practice, depression is diagnosed using clinical interviews and structured and semi-structured symptom severity scales (Beck's Depression Scale, American Psychiatric Association – DSM 5, ICD-10), all of which require accurate self-report from the patient. To the best of our knowledge, clinical professionals are not using any kind of quantification of EEG in the diagnostic process. Multiple factors (NESDA, 2018) make detection and diagnosis of depression often difficult. The use of biomarkers could facilitate diagnosis and potentially prevent future episodes, all of which garnered significant attention in the past decade. Most quantitative EEG (QEEG) measures, connectivity measures, vigilance-based measures, and sleep-related EEG measures use power spectrum analysis to quantify changes in brain activity related to depression. This approach repeatedly showed abnormal frontal asymmetry in alpha activity (Allen et al., 2004; Stewart et al., 2010) reduced slow-wave activity in sleep (Nissen et al., 2006), increased alpha activity (Kemp et al., 2010; Köhler et al., 2011; Baçar et al., 2011), and decrease in alpha synchronization in the right fronto-central and centro-parietal connections (Kim et al., 2013). Other studies focused on changes in theta (Knott et al., 2001; Ricardo-Garcell et al., 2009), and beta activity (Rohet al., 2016). Van der Vinne et al (2017) showed that frontal asymmetry although well established in the literature fail to serve as a prognostic biomarker for depression.

Among others, a recent study offered a new insight into the possible mechanisms behind major depressive disorder (MDD) (de Kwaasterniet et al., 2013). De Kwaasterniet and colleagues confirmed abnormal functional connectivity in the fronto-lymbic system by utilizing digital tractography imaging (DTI) and fMRI. A compromised second part of uncinate fasciculus in MDD seems to be correlated with increased functional connectivity but also with the severity of the disease. Kim et al (2013) showed by applying graph theory to EEG that 'functional topological architecture of the resting-state brain network is disrupted in bipolar disorder'. That study also revealed impaired neural synchronization at resting state as well as a disruption of functional connectivity. Both studies confirmed that the disturbance of white matter tracts is associated with abnormal functioning of fronto-lymbic system – a key indicator of depression. Having proven the efficacy of highly sophisticated methodologies, for the purpose of detection, we propose to investigate simple and low-cost EEG as a measure of electrical activity in the human brain which might reflect that deep brain regions' change on cortex. It is well described in the literature that a brain is a highly complex, nonlinear and mostly irregular system (Eke et al., 2002; Goldberger et al., 2001; Stam, 2005). There is increasing recent evidence that the use of non-linear methods may provide significant advantages in deciphering the physiological processes underlying EEG signals (Acharya et al., 2005; Liang et al., 2015; Stokić et al., 2015) and EEG remains the most accessible method (compared to others like fMRI which are far more expensive) for supporting decisions in clinical practice. EEG has the potential to be utilized as a screening method for a variety of psychiatric disorders.

There is a kind of a consensus between researchers that relying on one nonlinear measure might be misleading (Burns & Rajan, 2015), and there are also extremes in published studies with ten or more nonlinear measures applied (Liang et al, 2015), showing that every one of them is providing another kind of information about the signal under study. In our previous work, while testing the effectiveness of different combinations of nonlinear measures we found that Higuchi's fractal dimension (HFD) and sample entropy (SampEn) are particularly well matched in a methodological sense (in press) showing different sensitivity for frequency content of the signal. SampEn shows better performance in the lower frequency band, and HFD in higher frequencies of EEG. Both nonlinear measures are used to examine the complexity of signal: HFD is a complexity measure operating directly in time-domain, while SampEn is regularity statistics showing how predictable/irregular signal is. We then turned to establishing a methodology with measures which are computationally fast and robust to artifacts in the signal, and which could be clinically applicable. That methodology could be applied at different points in the diagnostic process to give the clinician additional support for a diagnostic decision. We did not attempt here to examine the correlation between previous medical histories of patients, but to show how proper quantification of their EEG can be used as an independent biomarker. Therefore, in this study, we tested the use of HFD and SampEn to discriminate the complexity of the brain's neuronal activity in patients diagnosed with depression.

'Data mining is the extraction of implicit, previously unknown, and potentially useful information from data' (Witten and Frank, 2005), and machine learning as a part of that discipline attracted a lot of attention lately. There is a relatively small number of publications dealing with the application of machine learning (data mining) algorithms to depression recognition using EEG. Ahmadlou, et al. (2012) compared two main algorithms for calculating fractal dimension from EEG. HFD was shown to provide better discrimination (91.3%) compared to Katz's Fractal Dimension (they used Enhanced Probabilistic Networks). Bachman et al (2013) used HFD together with SASI (a novel spectral measure) as a discriminator. They found an increased complexity of EEG in MDD. In their recent study (Bachmann et al., 2018) they also used Linear regression with HFD as feature with accuracy of 77%. Hosseinifard et al. (2013) used spectral and three nonlinear measures and found that classical spectral measures did not prove to be useful for classification. The aim of their study was to improve the accuracy of classifiers in the combination with different nonlinear measures. In addition, it has been found that Support Vector Machines (SVM) provided the best classification results compared to other methods such as Decision Tree (DT), k-Nearest Neighbor (kNN) and Naïve Bayes (NB) (Bairy et al., 2015). Recently published detailed mathematical description of interconnection of HFD and Fourier analysis (classical spectral analysis applied usually in EEG) components strongly suggest that the use of Higuchi's' Fractal Dimension and Fast Fourier Transform (FFT) is redundant because fractal dimensions are weighting functions of Fourier's amplitudes (Kalauzi et al., 2012). Complex signals are information-rich. They are also fractal in its nature for showing multiscalling and self-similarity. Another reason behind our preference to utilization of nonlinear measures in analysis of EEG is that nonlinear (nonstationary and irregular) signals 'defy comprehensive understanding by a classic reductionist approach' (Goldberger, 2006). Even the simplest nonlinear signals (originating from complex systems) will 'foil the criteria of proportionality and superposition characteristic for linear systems' (Goldberger et al., 2002; physionet.org). After performing complexity analyses we decided to compare seven different classifiers with a different combination of features and with a different number of principal components (PCs) from Principal Component Analysis (PCA) (Jolliffe, 2002). The aim of the study was to test the usefulness of employing nonlinear measures of complexity changes in EEG and machine learning to separate patients diagnosed with depression from healthy controls. We utilize seven different methods of classification (Logistic Regression, Support Vector Machines both with the linear and polynomial kernel, Multilayer Perceptron, Decision Tree, Random Forests, and Naïve Bayes). Our aim was to show that with properly selected nonlinear features and additional PCA processing every supervised learning method used can obtain high accuracy. The idea we want to convey here is that utilization of appropriate nonlinear measures to characterize an EEG signal is crucial for highly accurate classification; if features are appropriately generated, any classifier applied with those features is performing with high accuracy.

**MATERIALS AND METHODS**

Since the overall goal of the study was to demonstrate the usefulness of non-linear features for classification of patients diagnosed with depression based on EEG signals, we calculated Higuchi's Fractal Dimension (HFD) and Sample Entropy (SampEn) of EEG time series (series of data points indexed in time order from raw signal). Subsequently, we applied supervised machine learning algorithms to assess classification accuracy. To determine the linear dependence of EEG features, we utilized correlation analysis. We applied PCA to determine the influence of linear feature extraction on classification accuracy. Also, PCA is known from the literature for its possibility to reduce dimensionality of feature set making machine learning models more sensitive. We examined various classification algorithms, ranging from simple and linear to highly non-linear: Logistic Regression (LR), Support Vector Machines (SVM) both with the linear and polynomial kernel, Multilayer Perceptron (MP), Decision Tree (DT), Random Forests (RF) and Naïve Bayes (NB).

The non-linear features (both HFD and SampEn) were computed using a custom program written in Java. The classification was performed using Weka software (Weka v. 3.8, University of Waikato) (Hall et al., 2009). Principal component analysis was computed using Matlab (Matlab v. R2015b, Mathworks). Statistical analysis was performed using SPSS software (IBM SPSS Statistics 20).

**Participants**

The data used for this research were recorded at the Institute for Mental Health in Belgrade, Serbia. The subjects were 23 patients diagnosed with depression (13 women and 10 men), 24 to 68 years old (mean 31.53, SD 10.21). All of them were examined by senior clinical psychiatrist (the diagnosis was made according to the ICD-10 classification) and all were medicated. As a control we used the EEG records of 20 age-matched (mean 30.14, SD 8.94) healthy controls (10 males, 10 females) with no previous history of any neurological or psychiatric disorders, recorded at the Institute for Experimental Phonetic and Speech Pathology in Belgrade, Serbia. Healthy controls underwent general medical examination and testing with clinical psychologist to confirm their inclusion in the study. The participants from both groups were all right-handed, according to the Edinburgh Handedness Inventory (Oldfield, 1971). All the participants were informed about the experimental protocol and signed informed consent forms. The protocol was approved by the Ethics Committees of the participating institutions (Ethics Committee of the Institute for Mental Health, October 27th 2015, Approval number 30/59, and Ethics Committee of the Scientific Council of Institute for Experimental Phonetics and Speech Pathology, September 25th 2015, Approval number 87-EO/15). All procedures were in accordance with the ethical standards of the institutional and/or national research committee and with the 1964 Helsinki declaration and its later amendments or comparable ethical standards.

**Data acquisition**

The patients' EEGs were recorded after a visit to a recommended psychiatrist. EEG was recorded in the resting state with standard 10-20 system using NicoletOne Digital EEG Amplifier (VIAYSYS Healthcare Inc. NeuroCare Group), sitting upright in the comfortable chair, with closed eyes and without any stimulus. Both rooms were in Faraday's cage by design of building, noise was kept on both places below 42dB (measured using Phonometer), temperature was kept at 22 degrees Celsius, the light is a dimly daily light, the person was sitting surrounded by white curtains with some daily light through it. All the participants (from both groups) were recording between 10am and 12h (noon). EEGs were obtained from 19 electrodes in a monopolar montage with reference set to earlobes (Electro-cap International Inc. Eaton, OH USA) using sampling rate of 1 kHz and electrode resistance of less than 5 k$\Omega$. Bandpass was 0.5-70 Hz. For the control group, the same setup was used (10/20 system for electrode placement, the same electroconductive gel, resistance, calibration, etc.), but on the Nihon Kohden apparatus,

EEG 1200K Neurofax with Electrocap (model number 16755) International, Inc. Previous studies (for a detailed review see Pivick, 1993) compared results for the same subject on different equipment and concluded that intra-subject variability is small. We kept all the settings identical in all measurements.

Every recording lasted 3 minutes. Subjects were instructed to reduce any movement. We had to discard records of two patients from further analysis due to low voltage EEG in one male participant's case and epileptic seizure very close in time from the recording of one female participant. We used records from 21 patients diagnosed with depression and 20 age-matched healthy controls for this study. Artifacts were carefully inspected by two independent experts. The artefacts were not removed from the EEG signal. We analyzed only artefact free segments of the EEG trace, as we did not want to distort the signal by "fusing" selected parts (after artefact removal). From artifact-free traces we extracted three epochs for further analysis; every epoch was 5s (5000 samples) long. Altogether there were three epochs for every person recorded, resulting in 2337 epochs for further analysis.

**Fractal analysis**

Fractal dimension (FD) of a signal is a measure of its complexity and self-similarity in the time domain. FD is a number in the interval [1, 2]. Generally, higher self-similarity and complexity result in higher FD (Eke et al., 2002). The fractal dimension of EEG was calculated using Higuchi's algorithm (Higuchi, 1988) demonstrated to be the most appropriate for electrophysiological data (Esteller et al., 2001; Castiglioni, 2010). This method works directly in the time domain, gives a reasonable estimate of the fractal dimension even in the case of short signal segments and is computationally fast (since it does not attempt to reconstruct the strange attractor, like described in Stam, 2005). We computed the Higuchi algorithms with the maximal scale (Higuchi, 1988) $k_{max}$= 8 shown to perform the best for this type of signals (Spasić et al., 2005). Fractal dimensions were calculated for each electrode for the same duration of signal (the epoch of recorded EEG) for all the participants, and the calculated values formed ensembles (sets of variables) for further analysis. For calculating HFD we used in-house written script in Java programming language.

**Sample Entropy Analysis**

Another nonlinear measure, Sample Entropy (SampEn) was computed according to Richman and Moorman (2000). SampEn estimates signal complexity by computing the conditional probability that two sequences of a given length, *m*, similar for *m* points, remain similar within tolerance *r* at the next data point (when self-matches are not included). Mathematically, SampEn is the negative natural logarithm of the conditional probability that two sequences similar for *m* points remain similar at the next point. Thus, SampEn measures the irregularity of the data (the higher values, the less regular signal) that is related to signal complexity (Goldberger et al., 2002). Based on the changes of SampEn we can conclude in which direction the changes of the signal went (is it more or less complex). In accordance with a previous study (Molina-Picó et al., 2011), we used a tolerance level of $r = 0.15$ times the standard deviation of the time series (series of samples from raw recording EEG) and $m = 2$. For calculating SampEn we used in-house written script in Java programming language.

**Statistical analysis**

HFD resulted in 19 features (the number of electrodes when recording EEG), and SampEn also resulted in 19 features. We merged them to get 38 features for further (supervised) machine learning analysis. To determine whether the HFD and SampEn feature values significantly vary between EEG electrodes and between the groups (patients vs. control) we used MANOVA (SPSS Statistics version 20.0, SPSS Inc, USA), followed by posthoc Bonferroni tests (for comparison of each of 19 electrode's HFD and SampEn value between groups-HC and DP, resulting in 19 comparisons).

**Classifiers**

We compared the performance of several classifiers implemented in Weka software (Hall et al., 2009) with their default parameter values to discriminate between patients diagnosed with depression and controls. All classifiers are applied to normalized features. Normalization was performed by subtracting sample means and dividing by sample standard deviation such that the inputs of algorithms have zero means and unit standard deviation. To reduce the dimensionality of the feature set and decorrelate the features, we utilized PCA (Jolliffe, 2002) to obtain *m* principal components (PCs) corresponding to the largest eigenvalues of the sample covariance matrix. We defined the percentage of the explained variance by first *m* PCs as the ratio between sums of variances of *m* PCs and original variables.

*Naïve Bayes classifier* (John & Langley, 1995) is a maximum a posteriori classifier that outputs the class $c_i$ with the highest probability given the observed feature values. The posterior probability conditioned by the set of attributes $\{x_1, \ldots, x_k\}$ (assumed independent) is calculated using the Bayes' theorem (Bishop, 1995) as the product of posterior Gaussian probabilities (Witten & Frank, 2005).

*Logistic regression* (Cox, 1958) estimates the class conditional probability using a linear combination of features and logistic regression function. The model coefficients are estimated using the LogitBoost algorithm (Friedman et al., 2000).

*Multi-layer perceptron* (MLP) is a generalization of logistic regression with an additional processing layer. MLP estimates a class based on thresholding the value of its output processing unit (Friedman et al., 2000; Haykin, 2009). The output unit applies a non-linear transfer function to a linear combination of the outputs of hidden neurons. Each hidden neuron applies a transfer function to a linear combination of the inputs. We utilize MLPs with $\left\lceil \frac{k+1}{2} \right\rceil$ hidden neurons, back propagation algorithm with learning rate 0.3 and momentum 0.2 and 500 training epochs to determine the coefficients (Haykin, 2009).

*Support vector machines* (SVM) classify by partitioning a feature space by a decision boundary, linear in transformed space, defined by the kernel function, and uniquely determined by a subset of data – support vectors (Jolliffe, 2002). SVMs produce a maximal margin classifier that maximizes the distance between the decision boundary and the support vectors. In this study, we utilized linear and polynomial (quadratic) kernel functions (Jolliffe, 2002), soft-margin classifier with regularization constant *C*=1 and a sequential minimal optimization algorithm (Platt, 1998). SVMs by design, maximize classifier margin, and hence, probably, minimize overfitting.

*Decision trees* (Quinlan, 1993) recursively partition feature partition space in regions corresponding to classes by choosing a feature that provides the highest information gain. The partition stops when the minimal number of 2 samples per node of a decision tree is reached. In the pruning phase, based on the estimation of the classification error (using a confidence level here set to 0.25) the complexity of the model may be reduced and its generalization capacity thus improved (Vapnik, 1988).

*Random forests classifier* utilizes an ensemble of unpruned trees (Breiman, 2001). The classification is performed by combining classes predicted by ensemble members. Unlike C4.5 algorithm, in each node of a tree, a random subset of the features is considered for partitioning. In this study, we utilize ensembles with 100 members and consider int($\log_2 k$)+1 random features for each split.

**Evaluation of classifier's performance**

Classification accuracy was evaluated through a cross-validation procedure in which the dataset was split into K subsets of approximately equal size K-1 subsets were used to fit a classification model and the remaining subset to evaluate the classifier. This procedure was repeated K times such that a classifier was evaluated in each subset. In this study, we used K=10 (Picard & Cook, 1984; Kohavi, 1995). The classification accuracy was assessed through overall accuracy—the percentage of correctly classified samples—and using the area under the ROC curve (AUC). The overall accuracy of useful classifiers in two-class problems ranges from 50% to 100%. The ROC curve is created by plotting true positive rate (the proportion of samples with depression that are detected as such) vs. false positive rate (the ratio of the total number of controls incorrectly detected as with depression and the total number of controls). AUC ranges from 0.5 (for a classifier that randomly guesses a class) to 1 (for an ideal classifier) (Fawcett, 2006; Till & Hand, 2012).

In machine learning, model complexity (of the classifier not of the signal) is defined as ability of a classifier to distinguish among classes that are separated with multivarious surfaces. Here, model complexity was measured using the Vapnik-Chevronenkis (VC) dimension. Burges (1998) indicates that the models with a small number of parameters may have larger VC dimension and complexity. According to statistical learning theory (Vapnik, 1988), the classification accuracy on test data (measured by ten-fold cross-validation method in this study) decreases with a factor that is directly proportional to the VC dimension of the model and inversely proportional with the size of the training data set. Among the classification method considered in this paper, multilayer perceptron and decision tree have VC dimension that increases with the number of utilized features for classification.

**RESULTS**

**Fractal and SampEn analysis**

Our results showed that HFD for patients diagnosed with depression ranged from 1.0812 to 1.1553 and for controls from 1.0194 to 1.0198. SampEn of patients diagnosed with depression ranged from 0.3999 to 0.4160 and in the control group from 0.1417 to 0.1591. The first level of analysis was testing the existence of differences in HFD values and SampEn values between Patients (P) and Control (C) groups. MANOVA has utilized with factors Electrode (19 electrodes: Fp1, Fp2, F3, F4, C3, C4, P3, P4, O1, O2, F7, F8, T3, T4, T5, T6, Fz, Cz, Pz) and Group (P and C).

Results of HFD analysis (Fig. 1a) showed a statistically significant effect of group: $F(1, 570) = 159.965$, $p<0.001$, as well as interaction of group and electrode: $F(18,570) = 1.677$, $p= 0.039$ on values of calculated HFD.

(Figure 1 about here)

There is no statistically significant difference between electrodes inside the P and C group. Our results show higher values of HFD in P group compared to C group. A significant difference in HFD values exists between P and C group for every electrode (post hoc Bonferroni correction, p<0.05), except for electrodes P3 and F4.

Results of SampEn analysis (Fig. 1b) showed a statistically significant effect of group:

F(1, 570) = 625,914, *p*<0.001 on SampEn values. Results showed higher values of SampEn of EEG from P group when compared to C group. There were no statistically significant differences among electrodes inside the P and C group when tested independently (post hoc Bonferroni correction, p<0.05). A significant difference between P and C group in SampEn values exists for every electrode (post hoc Bonferroni correction, p ≤ 0.001). Subjects from P group have higher values of SampEn calculated from EEG recorded on all the electrodes when compared to C group values.

**Correlation of HFD and SampEn values**

Figure 2 contains calculated Pearson's correlation coefficients (Devore, 2012) between the features. Figure 2a shows correlation coefficients between HFD values calculated for 19 electrodes. The minimal and the maximal correlation coefficient values were 0.77 and 0.99, respectively. Figure 2b displays the correlation coefficient values between SampEn features. The minimal and maximal values of the correlation were respectively 0.86 and 0.99. Figure 2c depicts the correlation between pairs of SampEn and HFD features; the minimal and maximal correlation coefficients were 0.44 and 0.79. All the estimated values of correlation coefficients were significantly different from 0 (p<0.005).

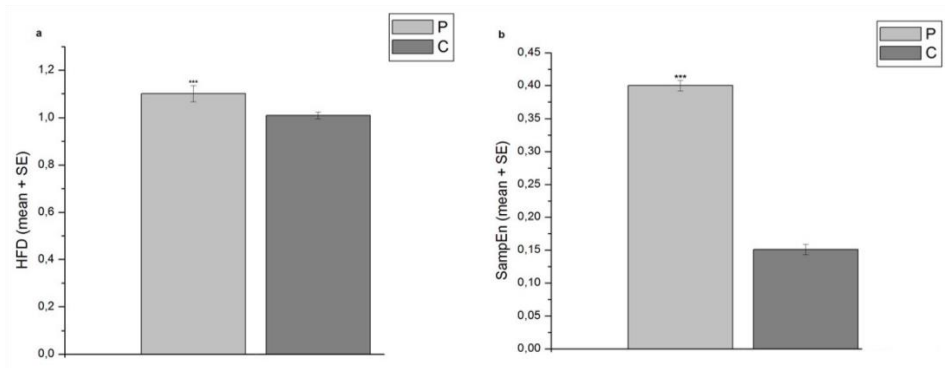

**Figure 1. Differences between Patients (P) and Control (C) in HFD (a) and SampEn (b) values (all electrodes averaged).** For both HFD and SampEn mean values + standard error are presented. *** ($p < 0.001$), HFD (Higuchi's fractal dimension), SampEn (Sample Entropy).

When two feature sets are less correlated, but each of them separately provides good classification accuracy, then the addition of one feature set to another in principle could results in a higher classification accuracy than when the feature sets are used separately. In such a case, we could benefit from (relative) orthogonality of two informative feature sets.

**Classification**

Table 1 shows classification results with different classifiers: multilayer perceptron, logistic regression, Support Vector Machines (SVM) with the linear and polynomial kernel ($p=2$), Decision Tree, Random Forest, and Naïve Bayes. Accuracy and area under the ROC curve (AUC) are shown for three different sets of features: HFD, SampEn, and their combination. Also, the average accuracies for all classifiers are shown on each feature set and for each method.

**Table 1**. Classification results for different classifiers and three different sets of features.

| | Features | | | | | | Average accuracy per classifier |
|---|---|---|---|---|---|---|---|
| | HFD | | SampEn | | SampEn+HFD | | |
| Classifier | Accuracy | *AUC* | Accuracy | *AUC* | Accuracy | *AUC* | |
| Multilayer perceptron | 100% | *0.998* | 97.56% | *0.995* | 95.12% | *0.995* | 97.56% |
| Logistic regression | 92.68% | *0.960* | 92.68% | *0.995* | 97.56% | *0.998* | 94.31% |
| SVM with linear kernel | 85.37% | *0.857* | 95.12% | *0.952* | 95.12% | *0.952* | 91.87% |
| SVM with polynomial (quadratic) kernel (p=2) | 80.49% | *0.810* | 95.12% | *0.952* | 95.12% | *0.954* | 90.24% |

| Decision tree | 92.68% | 0.904 | 97.56% | 0.975 | 95.12% | 0.952 | 95.12% |
| Random forest | 92.68% | 0.970 | 95.12% | 0.988 | 92.68% | 0.987 | 93.49% |
| Naïve Bayes | 85.37% | 0.945 | 92.68% | 0.990 | 92.68% | 0.983 | 90.24% |
| **Average accuracy per feature set** | **89.90%** | | **95.12%** | | **94.77%** | | |

*Note* – HFD (Higuchi's fractal dimension), SampEn (Sample entropy), SVM (Support Vector Machines), AUC (area under the curve; related to Receiver Operating Characteristic – ROC curves).

Figure 3 shows the percentage of explained variance of a set of HFD and SampEn features as a function of the number of principal components. Fig. 4 shows absolute values of loads used to calculate first ten principal components from HF and SampEn features. Table 2 shows classification results of different classifiers that use various numbers of principal components. The principal components were computed on a dataset containing HFD and SampEn features and were normalized to have zero mean and unit standard deviation. The variance of the features explained by the corresponding principal components is also shown.

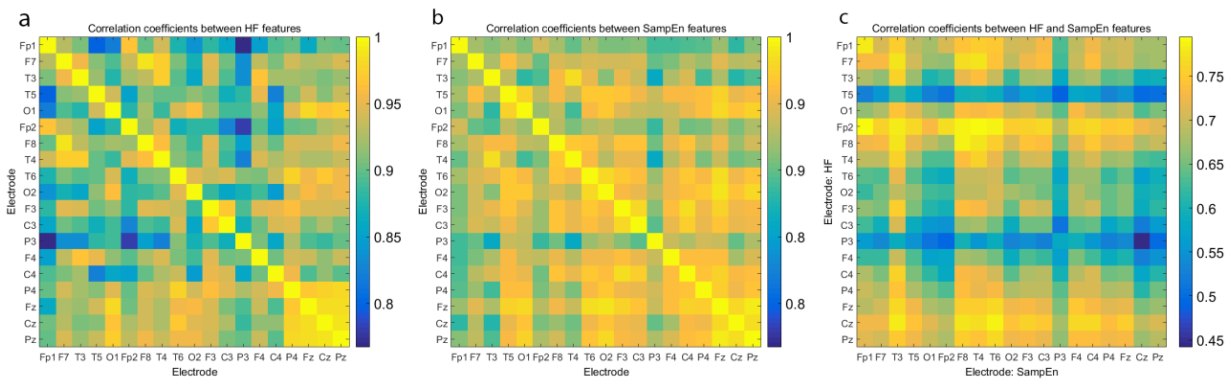

**Figure 2. Pearson's correlation coefficients computed between pairs of features.** The value of correlation coefficient is color-coded (the scales at each figure are different). The labels on x and y axis denote the electrode corresponding to a feature; a) Correlation coefficients between pairs of HFD features; b) Correlation coefficients between pairs of SampEn features; c) Correlation coefficients between SampEn (on x axis) and HFD (on y axis) feature.

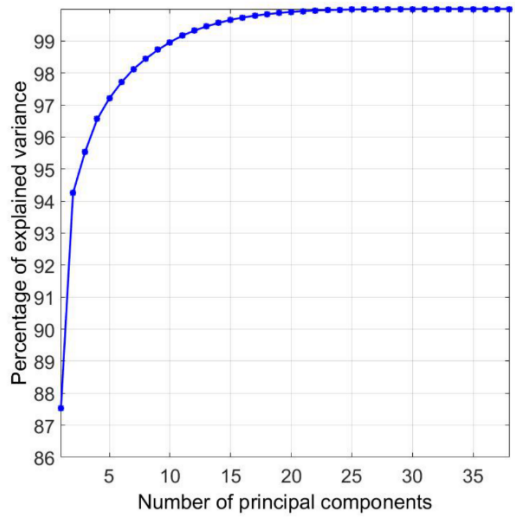

**Figure 3. Percentage of explained variance vs. number of principal components of HFD and SampEn features.** For each number of principal components *m* we calculated the percentage of explained variance. When the number of principal components reaches 20, the percentage of explained variance saturates to a value close to 100%.

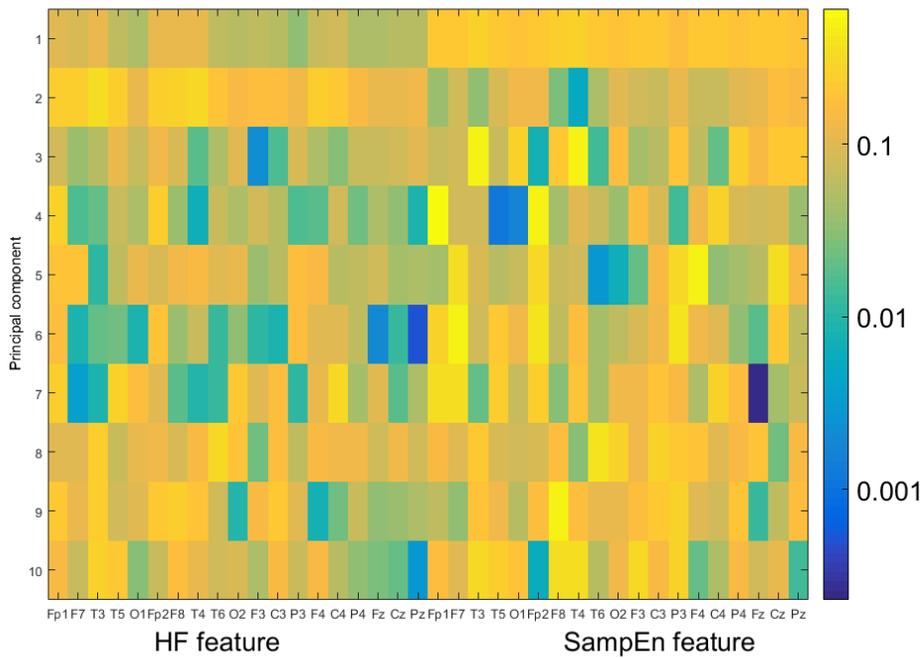

**Figure 4. Absolute values of principal components loads for first 10 principal components.** Each row contains indicates the coefficients multiplying corresponding non-linear feature in order to generate a principal component.

**Table 2**. Classification results for different classifiers and different number of principal components of the features from SampEn and HFD sets

| Number of principal components | 1 | | 2 | | 3 | | 10 | |
|---|---|---|---|---|---|---|---|---|
| Explained variance | 87.53% | | 94.25% | | 95.54% | | 98.86% | |
| Classifier | Accuracy | AUC | Accuracy | AUC | Accuracy | AUC | Accuracy | AUC |
| *Multilayer perceptron* | 92.68% | *0.983* | 92.68% | *0.943* | 92.68% | *0.950* | 95.12% | *0.994* |
| *Logistic regression* | 90.24% | *0.981* | 90.24% | *0.950* | 90.24% | *0.945* | 95.12% | *0.929* |
| *SVM with linear kernel* | 85.37% | *0.857* | 82.92% | *0.833* | 85.37% | *0.857* | 90.24% | *0.905* |
| *SVM with polynomial (quadratic) kernel (p=2)* | 73.17% | *0.738* | 68.29% | *0.690* | 85.37% | *0.857* | 95.12% | *0.952* |
| *Decision tree* | 92.68% | *0.894* | 92.68% | *0.894* | 90.24% | *0.933* | 90.24% | *0.933* |
| *Random forest* | 87.80% | *0.981* | 87.80% | *0.981* | 95.12% | *0.987* | 95.12% | *0.996* |
| *Naïve Bayes* | 95.12% | *0.983* | 97.56% | *0.981* | 95.12% | *0.988* | 95.12% | *0.988* |
| *Average accuracy* | 88.15% | | 87.45% | | 90.59% | | 93.73% | |

*Note* – SVM (Support Vector Machines), AUC (area under the curve; related to Receiver Operating Characteristic – ROC curves)

Figure 3 and Table 2 indicate that the large portion of the variance of combined HFD and SampEn features can be explained by a small number (Pokrajac et al., 2014) of principal components. e.g., the first principal component explains 87.53% of the variance; first three components explain more than 95% and the first 10 components close to 99% of the variance. Using only the first principal component, it is possible to achieve a classification accuracy of up to 95.12% (Table 2). The best performance was achieved using the Naïve Bayes method (in this case, when only one feature is utilized, the assumption of feature independence is automatically satisfied). The classification accuracy generally increases with the number of principal components used as classifiers' inputs (the average accuracy of all classifiers is 88.15% with 1 and 93.73% with 10 principal components used).

**DISCUSSION**

The major finding of this study is that the extraction of non-linear features linked to the complexity of EEG signals can lead to high and potentially useful separation between signals taken from control subjects and patients diagnosed with depression. Specifically, we demonstrated that Higuchi's Fractal Dimension (HFD) and Sample Entropy (SampEn) could be used as suitable features for various machine learning classification techniques. In other words, proper choice of a non-linear feature extraction method (HFD/SampEn) simplifies an important classification problem and makes it tractable. To the best of our knowledge, we were the first to apply this specific feature extraction method on this particular classification task.

When compared to the present literature our research has substantial originality, including: combined use of HFD and SampEn on broadband EEG signal (with minimal pre-processing); a variety of classification methods applied and demonstrated robustness on choice of classification method when non-linear features are utilized and application of principal component analysis (PCA) and demonstration of their power for feature extraction. The rest of this Discussion concentrates on these aspects of our work.

In the present literature, only a few studies applied an approach similar to ours (Bachman et al., 2018; Ahmadlou et al, 2012; Hosseinifard et al 2013; Acharya et al., 2015). Fractal dimension (Ahmadlou et al., 2012) and both linear and nonlinear measures of EEG (Hosseinifard et al., 2013) were applied to the classification of patients diagnosed with depression and healthy controls. It was found that nonlinear features gave better results, when compared to spectral ones, in the classification of patients diagnosed with depression (Hosseinifard et al., 2013). Note that the use of reductionistic approaches, such as Fourier's analysis, was found inferior (Rabinovich, 2006; Klonowski, 2007). The rationale here is a part of Complexity theory which led to consensus among researchers dealing with nonlinear analysis; key properties of linear systems are proportionality and superposition (Goldberger, 2006). Nonlinear systems defy comprehensive understanding by a classic reductionist approach (like Fourier's analysis), since they do not obey proportionality and superposition. Since human brain is one of the most complex systems we know of, analyzing the signal originating from it (EEG) by utilization of reductionistic method could be misleading.

In line with previous findings (Bachmann et al, 2013; Ahmadlou et al., 2012) about measures used for characterization of EEG, in our study HFD detected increased complexity of EEG recorded from patients diagnosed with depression in comparison to healthy controls. This is also in agreement with a study by Hosseinifard et al. (2013) who demonstrated that non-linear features, such as HFD, correlation dimension, and Lyapunov coefficient are more discriminative than linear features. The main difference with our work is that we analyzed broad band signal, and other divided the signal on standard spectral bands. Ahmadlou et al. (2012) reported classification accuracy of 91.3% when using two differently calculated fractal dimension algorithms (Higuchi and Katz) as features and enhanced probabilistic neural networks for classification. Our results are in qualitative agreement with this finding. In addition to confirming previous results, we not only showed that SampEn can also effectively discriminates these two categories of EEG signals, but also it could have performance superior to HFD, see Table 1.

A variety of classification methods have been used in domains similar to ours: support vector machines (SVM), linear regression (LR), linear discriminant analysis (LDA), k-nearest neighbors (kNN), enhanced probabilistic neural networks (Ahmadlou et al., 2012; Hosseinifard et al., 2013; Acharya et al., 2015). The choice of a specific classification method is frequently a matter of bias of researchers (Pokrajac, 2014). Moreover, in absence of standardized data repositories, that exist in other domains (Lichman, 2013) and a strict statistical test for comparison of, frequently non-linear and non-parametric, classifiers (Efron, Tibshirani, 2007) direct comparison of accuracies among different methods and results from different publications is challenging.

Instead of attempting to compare classification of classifiers, our goal was to demonstrate that the usage of non-linear features can result in high classification accuracy regardless of a classifier choice. Observe that a similar methodological approach to validate the usefulness of feature extraction was taken in Pokrajac et al. (2014) in another domain. For this reason, we did not try to optimize classifier parameters, but utilized their default values (similar as in Unnikrishnan et al., 2016). The reported average accuracy of all the methods, as an indicator of the quality and usability of the features, was 95.12% on SampEn features (and 89.90% on HFD), see Table 1. To estimate accuracy of each particular method, we utilized a standard 10-fold cross-validation technique (Devijver, Kittler, 1982). Note that it resulted in qualitatively similar results as a bootstrapping technique applied by Ahmadlou et al. (2012).

Note that we examined classification methods with a range of underlying paradigms and complexity; the methods belong to statistics and supervised machine learning. Even the simplest methods, such as logistic regression (widely accepted in the medical community although not as a classification method in the strict sense) provided excellent classification accuracy. In fact, high classification accuracy of methods such as SVM with linear kernel indicate that, after a non-linear transformation, the data may become close to linearly separable; we are however aware that this may be specific for a particular dataset and should be tested on further data. Consistent with known properties of the Naïve Bayes classifier (Witten & Frank, 2005; Mitchell, 1997), it had good accuracy (e.g., 92.68%, AUC of 0.983 when applied on SampEn and HFD features combined) albeit the underlying assumption about feature independence is not satisfied. Presumably, due to the high correlation of features (see Fig. 2), there was no benefit of using random forests in comparison to standard decision tree classifiers.

Our results indicate that the use of SampEn features may result in classification results comparable to or better than HFD. Five out of seven examined classifiers provided better accuracy while six provided a higher area under ROC curve when applied on SampEn features. The accuracy of a linear model (SVM with the linear kernel) increased by almost 10% when applied to SampEn features. Similarly, the accuracy of SVM with the polynomial kernel increased by almost 15%. The relatively low accuracy of polynomial SVMs is in agreement with previously reported results (Hosseinifard et al., 2013).

To the best of our knowledge, there are no previous attempts to utilize a combination of HFD and SampEn features. In our experiment, the use of an augmented feature set consisting of HFD features and SampEn features did not lead to further improvement of accuracy for the majority of attempted classifiers (Table 1). Note that HFD features are less correlated to SampEn features, then HFD features or SampEn features are correlated among themselves. Namely, the maximal correlation between one HFD and a SampEn feature is 0.79. In contrast, the maximal correlation between two SampEn features or between two HFD features is larger than 0.98. The combination of two relatively uncorrelated features, such as HFD features and SampEn features, provides an opportunity for training of more expressive classification models that could result in better classification accuracy. This would fully exploit orthogonality of features when using more complex classification models that would generalize better if trained on larger datasets (Kecman, 2001). However, the prerequisite for achieving such increased accuracy is large enough size of the data set, which may be produced in follow-up studies. If the available data set is not sufficiently large, some machine learning algorithms suffer from potential of overfitting— when a learning algorithm, in attempts to minimize error on a training set, results in a model that has poor generalization abilities (Vapnik, 1998). Our results suggest this may be a case with random forests and multilayer perceptron, where the accuracy achieved with the combined (HFD+SampEn) feature set is smaller than using HFD or SampEn separately (Table 2). In contrast, machine learning models that have small or controlled complexity (e.g., expressed through VC dimension (Vapnik, 1998)), such as support vector machines, Naïve Bayes or logistic regression, did not express this behavior; in this model, the use of the augmented data set led to the same or increased accuracy.

Principal component analysis, a feature extraction method where a linear transformation on the original feature vector is applied to reduce its dimensionality (Joliffe, 2002) is applied in this paper in order to demonstrate that accurate classification is possible using a small number of principal components. Note this technique is typically utilized to decorrelate features, such as HFD and SampEn in our case (see Figure 2).

Using only the first principal component, it is possible to achieve a classification accuracy of up to 95.12%, using the Naïve Bayes method (Table 2). The classification accuracy generally increases with the number of principal components used as classifiers' inputs (the average accuracy of all classifiers is 88.15% with 1 and 93.73% with 10 principal components used). Since the data are close to linearly separable, SVM with linear kernel resulted in relatively high accuracy (85.37%) when only one PC is used. The accuracy further increased to 90.24% with 10 PCs. This can be in part explained by Cover's theorem (Witten, 2005) that data in higher dimensional spaces tend to be more linearly separable. The random forest method benefited from a larger number of utilized principal components since the method is based on randomly choosing one from a set of available features to split a decision tree node (when the number of PCs used is small, the set of available features is small).

The determination of the minimal number of PCs depends on the desired classification accuracy and is generally problem dependent (Pokrajac et al, 2014; Kecman, 2001). In this specific case, if a minimal AUC is set to 0.85 (corresponding to diagnostic tests considered good) (the source that was used to generate http://gim.unmc.edu/dxtests/Default.htm), three PCs are sufficient, according to Table 2.

Our results suggest that classifiers that could be implemented on a simple and inexpensive hardware and embedded to existing EEG devices. This itself may ultimately lead to potential everyday clinical usage of our methodology for providing computer-aided diagnostics of depression. The technology could be of interest, e.g., when burnout or extreme amount of stress, using the current diagnostic methods, are mistaken as symptoms of depression. Other researchers are testing whether similar methodology can help identify a patient as a good responder to particular treatment be it medication or transcranial direct electrical stimulation (Shahaf et al., 2017; Al- Kaysi et al., 2017).

Note that the values PCA loads (used to weight non-linear features in order to compute principal components), Fig. 4, indicate that all non-linear features contribute to calculated principal components. In other words, the use of PCA implies that the information is not contained in a signal from a particular electrode, but distributed through multiple electrodes. Therefore, the use of multiple electrode signals can contribute to better distinction between controls and patients diagnosed with depression. This is in agreement of previous fMRI and DTI findings (de Kwaasterniet et al, 2013). Namely, in patients diagnosed with depression, a decreased functional connectivity within fronto-limbic network and anatomical difference in second part of uncinate fasciculus—deep white matter tract connecting prefrontal cortices with limbic system is observed. Our hypothesis was that such dysfunction might result in compensation which might be detected on surface (cortex). The brain compensation in turns translates to an alleviated excitability on cortex, which can be observed on signals from multiple EEG electrodes. Note that in available literature, the number of utilized electrodes is smaller than in our study. Ahmadlou et al. (2012) used seven electrodes (prefrontal), while Bachmann et al. (2013) recommended two electrodes or one electrode (Bachmann et al, 2018).

Finally, we would like to emphasize that an extension of the method on larger data sets is needed prior to making a final conclusion about class separability and the potential applicability of the classification techniques for diagnostic purposes.

## CONCLUSION

In this study, we demonstrated that Higuchi's Fractal Dimension and Sample Entropy are capable of distinguishing between participants diagnosed with depression and healthy controls' EEG. If a feature extraction method results in good classification accuracy regardless of applied machine learning technique, this provides the evidence that the feature extraction method is useful. These results encourage further investigation with larger sample sizes towards potential diagnostic application in clinical medicine and psychiatry.

## REFERENCES

Acharya UR, Sudarshan VK, Adeli H, Santhosh J, Koh JE, Puthankatti SD, Adeli A. 2015. A Novel Depression Diagnosis Index Using Nonlinear Features in EEG Signals. *European Neurology* **74(1–2)**: 79–83. DOI: 10.1159/000438457.

Acharya UR, Faust O, Kannathal N, Chua T, Laxminarayan S. 2005. Nonlinear analysis of EEG signal at various sleep stages. *Computer Methods and Programs in Biomedicine* **80(1)**: 37–45. DOI:10.1016/j.cmpb.2005.06.011.

Ahmadlou M, Adeli H, Adeli A. 2012. Fractal analysis of frontal brain in Major depressive disorder. *International Journal of Psychophysiology* **8(2)**: 206–211. DOI: 10.1016/j.ijpsycho.2012.05.001.

Al –Kayasi AM, Al-Ani A, Loo CK, Powell TY, Martin DM, Breakspear M, Boonstra TW. 2017. Predicting tDCS treatment outcomes of patients with major depressive disorder using automated EEG classification. *Journal of Affective Disorders* **208**: 597-603.

Allen JJB, Urry HL, Hitt SK, Coan JA. 2004. The stability of resting frontal electroencephalographic asymmetry in depression. *Psychophysiology* **41:** 269–280. DOI:10.1111/j.1469-8986.2003.00149.x

American Psychiatric Association. 2013. Diagnostic and statistical manual of mental disorders (5th ed.). Washington, DC.

Arnone D, McIntosh AM, Ebmeier KP, Munafò MR, Anderson IM. 2012. Magnetic resonance imaging studies in unipolar depression: systematic review and meta-regression analyses. *European Neuropsychopharmacology* **22:** 1–16. DOI: 10.1016/j.euroneuro.2011.05.003.

Arnone D, McKie S, Elliott R, Juhasz G, Thomas EJ, Downey D, Anderson IM. 2013. State-dependent changes in hippocampal grey matter in depression. *Molecular Psychiatry* **18**: 1265–1272. DOI: 10.1038/mp.2012.150.

Bachmann M, Lass J, Suhhova A, Hinrikus H. 2013. Spectral asymmetry and Higuchi's fractal Dimension of Depression electroencephalogram. *Computational and Mathematical Methods in Medicine* 251638, DOI: 10.1155/2013/251638.

Bachmann M, Päeske L, Kalev K, Aarma K, Lehtmetmets A, Ööpik P, Lass J, Hinrikus H. 2018. Methods for classifying depression in single channel EEG using linear and nonlinear signal analysis. *Computer Methods and Programs in Biomedicine* **155**: 11-17.

Baggio PS, Ferucci R, Rigonatti SP, Covre P, Nitsche M, Pascual-Leone A, Fregni F. 2006. Effects of transcranial direct current stimulation on working memory of patients with Parkinson's disease. *Journal of Neurophysiological Sciences* **249**: 31-38. DOI: 10.1016/j.jns.2006.05.062.

Bairy GM, Bhat S, Eugene LWJ, Niranjan UC, Puthankatti SD, Joseph PK. 2015. Automated Classification of Depression Electroencephalographic Signals Using Discrete Cosine Transform and Nonlinear Dynamics. *Journal of Medical Imaging and Health Informatics* **5(3)**: 1–6. DOI: 10.1166/jmihi.2015.1418.

Basar E, Guntekin B, Atagun I. 2011. Brain's alpha activity is highly reduced in euthymic bipolar disorder patients. *Cognitive Neurodynamics* DOI: 10.1007/s11571-011-9172-y.

Beck AT, Steer RA, Ball R, Ranieri W. 1996. Comparison of Beck Depression Inventories-IA and –II in psychiatric outpatients. *Journal of Personality Assessment* **67(3)**: 588-597.

Bishop C. 1995. *Neural Networks for pattern recognition*. Oxford University Press, pp. 116–160.

Breiman L. 2001. Random Forests. *Machine Learning* **45(1)**: 5–32.

Burges CJC. 1998. A tutorial for support vector machines for pattern recognition. *Data Mining and Knowledge Discovery 2*, Kluwer Academic Publishers, Boston, pp. 121–167.

Burns T, Ramesh, R. 2015. Combining complexity measures of EEG data: multiplying measures reveal previously hidden information. *F1000Research* **4**:137. DOI: 10.12688/f1000research.6590.1.

Castiglioni P. 2010. What is wrong with Katz's method? Comments on: a note on fractal dimensions of biomedical waveforms. *Computers in Biology and Medicine* **40(11-12):** 950-952. DOI: DOI:10.1016/j.compbiomed.2010.10.001.

Cox DR. 1958. The regression analysis of binary sequences (with discussion). *Journal of the Royal Statistical Society B* **20**: 215–242.

Čukić M, Platiša M, Kalauzi A, Oommen J, Ljubisavljević MR. 2017. The comparison of Higuchi's fractal dimension and Sample Entropy analysis of sEMG: effects of muscle contraction intensity and TMS. *Fractal Geometry and Nonlinear analysis in Medicine and Biology*, Vol 3, Issue 2, 2017.

Davidson R. 2004. What does the prefrontal cortex 'do' in affect: Perspectives on frontal EEG asymmetry research. *Biological Psychology* **67**: 219–33. DOI: 10.1016/j.biopsycho.2004.03.008.

Devijver PA, Kittler J. 1982. Pattern Recognition: A Statistical Approach. London, GB: Prentice-Hall.

Devore JL. 2012. *Probability and Statistics for Engineering and the Sciences, 8th edn*. Brooks/Cole.

Efron B, Tibshirani R. 1997. Improvements on cross-validation: The .632 + Bootstrap Method. *Journal of the American Statistical Association* **92(438)**: 548–560. doi:10.2307/2965703.

Eke A, Herman P, Koscis L, Kozak LR. 2002. Fractal characterization of complexity in temporal physiological signals. *Physiological Measurements* **23(1)**: R1-R38.

Fawcett T. 2006. An introduction to ROC analysis. *Pattern Recognition Letters* **27**: 861–874. DOI: 10.1016/j.patrec.2005.10.010.

Friedman JH, Hastie T, Tibshirani R. 2000. Additive logistic regression: a statistical view of boosting (with discussion). *The Annals of Statistics* **28**: 337–407. DOI: 10.1214/aos/1016218223.

Goldberger AL, Peng CK, Lipsitz LA. 2002. What is physiologic complexity and how does it change with aging and disease? *Neurobiology of Aging* **23**: 23-26.

Hall M, Frank E, Holmes G, Pfahringer B, Reutemann P, Witten IH. 2009. The WEKA Data Mining Software: An Update. *ACM SIGKDD Explorations Newsletter* **11(1)**: 10–18. DOI:10.1145/1656274.1656278.

Hand RJA, Till DJ. 2012. Simple generalization of the area under the ROC curve for multiple class classification problems. *Machine Learning* **45**: 171–186. DOI: 10.1023/A:101092081.

Haykin S. *Neural Networks and Learning Machines* (3rd ed), Pearson Education. pp 122–218.

Henriques JB, Davidson RJ. 1991. Left frontal hypoactivation in depression. *Journal of Abnormal Psychology* **100**: 535–545.

Higuchi T. 1988. Approach to an irregular time series on the basis of the fractal theory. *Physica D* **31**: 277–283. DOI: 10.1016/0167-2789(88)90081-4.

Hogan MJ, Kilmartin L, Keane M, Collins P, Staff RT, Kaiser J, Lai R, Upton N. 2012. Electrophysiological entropy in younger adults, older controls and older cognitively declined adults. *Brain Research* **1445**: 1–10.DOI:10.1016/j.brainres.2012.01.027.

Hosseinifard B, Moradi MH, Rostami R. 2013.Classifying depression patients and normal subjects using machine learning techniques and nonlinear features from EEG signal. *Computer Methods and Programs in Biomedicine* **109(3)**: 339–345. DOI: 10.1016/j.cmpb.2012.10.008.

Jolliffe JL. 2002. *Principal Component Analysis* (2$^{nd}$ed), Springer –Verlag, pp. 10–150.

John GH, Langley P. 1995. Estimating Continuous Distributions in Bayesian Classifiers. In: *Proceedings of the Eleventh Conference on Uncertainty in Artificial Intelligence*, San Mateo, 338–345.

Kalauzi A, Bojic T, Vuckovic A. 2012. Modeling the relationship between Higuchi's fractal dimension and Fourier spectra of physiological signals. *Medical & Biological Engineering & Computing* **50(7)**: 689–699. DOI: 10.1007/s11517-012-0913-9.

Kecman V. 2001. *Learning and Soft Computing: Support Vector Machines, Neural Networks and Fazzy logic Models*. A Bradford Book, The MIT press, Cambridge, Massachusetts, London.

Kemp AH, Griffiths K, Felgham KL, Shankman SA, Drinkenburg W, Arns M, Clark CR, Bryant RA. 2010. Disorder specificity despite comorbidity: resting EEG alpha asymmetry in major depressive disorder and post-traumatic stress disorder. *Biological Psychology* **85(2)**: 350–354. DOI: 10.1016/j.biopsycho.2010.08.001.

Kim D, Bolbecker AR, Howell J, Rass O, Sporns O, Hetrick WP, Breier A, O'Donnell BF 2013. Disturbed resting state EEG synchronization in bipolar disorder: a graph-theoretic analysis. *NeuroImage: Clinical* **2**: 414–423. DOI: 10.1016/j.nicl.2013.03.007.

Klonowski W. 2007. From conformons to human brains: an informal overview of nonlinear dynamics and its applications in biomedicine. *Nonlinear Biomedical Physics* **1(1)**: 5. DOI:10.1186/1753-4631-1-5.

Knott V, Mahoney C, Kennedy S, Evans K. 2000. Pre-treatment EEG and its relationship to depression severity and paroxetine treatment outcome. *Pharmacopsychiatry* **33**: 201–205. DOI: 10.1055/s-2000-8356.

Kohavi RA. 1995. A study of cross-validation and bootstrap for accuracy estimation and model selection. *Proceedings of the Fourteenth International Joint Conference on Artificial Intelligence*. San Mateo, CA: Morgan Kaufmann **2,** 1137–1143.

Köhler S, Ashton CH, Marsh R, Thomas AJ, Barnett NA, O'Brien JT. 2011. Electrophysiological changes in late life depression and their relation to structural brain changes. *International Psychogeriatrics* **23**(**1**): 141–148.

Koolschijn PC, van Haren NE, Lensvelt-Mulders GJ, Hulshoff Pol HE, Kahn RS. 2009. Brain volume abnormalities in major depressive disorder: a meta-analysis of magnetic resonance imaging studies. *Human Brain Mapping* **30**: 3719–3735. DOI: 10.1002/hbm.20801.

Kwaasteniet BD, Ruhe E, Caan M, Rive M, Olabarriaga S, Groefsema M. 2013. Relation Between Structural and Functional Connectivity in Major Depressive Disorder. *Biological Psychiatry* **74**: 40–47. DOI: 10.1016/j.biopsych.2012.12.024.

Liang Z, Wang Y, Sun X, Li D, Voss LJ, Sleigh JW, Hagihira S, Li X. 2015. EEG entropy measures in anesthesia, Frontiers in Computational Neuroscience **9**:16. doi: 10.3389/fncom.2015.00016.

Lichman M. 2013. UCI Machine Learning Repository [http://archive.ics.uci.edu/ml]. Irvine, CA: University of California, School of Information and Computer Science.

Liu H, Matoda H. (editors), Computational Methods of Feature Selection, Chapman and Hall/CRC, 2008.

Mathews CD, Loncar D. 2006. Projections of Global Mortality and Burden of Disease from 2002 to 2030. *PlosMedicine* **3(11)**: e442. DOI:10.1371/journal.pmed.0030442.

Mitchell TM. 1997. *Machine Learning,* McGraw-Hill. Pp. 177-180.

Molina-Picó A, Cuesta-Frau D, Aboy M, Crespo C, Miró-Martínez P, Oltra-Crespo S. 2011. Comparative study of approximate entropy and sample entropy robustness to spikes. *Artificial Intelligence in Medicine* **53(2)**: 97–106. DOI: 10.1016/j.artmed.2011.06.007.

Murray CJ, Vos T, Lozano R, Naghavi M, Flaxman AD, Michaud C, Ezzati M, Shibuya K, Salomon JA, Abdalla S. 2013. Disability-adjusted life years (DALYs) for 291 diseases and injuries in 21 regions, 1990-2010: a systematic analysis for the Global Burden of Disease Study 2010. *The Lancet* **380**:2197–2223. DOI 10.1016/S0140-6736(12)61689-4.

Nissen C, Feige B, Nofzinger EA, Voderholzer U, Berger M, Riemann D. 2006. EEG slow wave activity regulation in major depression. *Somnologie* **10**: 36–42. DOI:10.1111/j.1439-054X.2006.00083.x

Netherlands Study of Depression and anxiety - NESDA 2018. http://www.emgo.nl/research/international-collaborations/longitudinal-cohort-studies/netherlands-study-of-depression-and-anxiety

Oldfield RC. 1971. The assessment and analysis of handedness: the Edinburgh inventory. *Neuropsychologia* **9**: 97–113.

Picard R, Cook D. 1984. Cross-Validation of Regression Models. *Journal of the American Statistical Assoc*iation **79**: 575–583. DOI: 10.1080/01621459.1984.10478083.

Pivick RT, Broughton RJ, Copola R, Davidson EJ, Fox N, Nuwer MR. 1993. Guidelines for the recording and quantitative analysis of encephalographic activity in research contexts. *Psychophysiology* **30(6)**: 547–558. DOI: 10.1111/j.1469-8986.1993.tb02081.x

Platt J. 1998. Fast Training of Support Vector Machines using Sequential Minimal Optimization. In *Advances in Kernel Methods - Support Vector Learning (Eds. Schölkopf B, Burges CJC),* MIT Press, pp. 41–65.

Pokrajac D, Lazarevic A, Kecman V, Marcano A, Markushin Y, Vance T, Reljin N, McDaniel S, Melikechi N. 2014. Automatic classification of laser-induced breakdown spectroscopy (LIBS) data of protein biomarker solutions. *Applied Spectroscopy* **68(9)**: 1067–1075. DOI 10.1366/14-07488.

Quinlan R. 1993. *Programs for Machine Learning*, Morgan Kaufmann Publishers, pp. 17–45.

Ricardo-Garcell J, Gonzalez-Olvera JJ, Miranda E, Harmony T, Reyes E, Almeida L, Galan L, Diaz D, Ramirez L, Fernandez-Rouzas A, Aubert E. 2009. EEG sources in a group of patients with major depressive disorders. *International Journal of Psychophysiology* **71(1)**: 70–74. DOI: 10.1016/j.ijpsycho.2008.07.021.

Richman JS, Moorman JR. 2000. Physiological time-series analysis using approximate entropy and sample entropy. *American Journal of Physiology: Heart and Circulatory Physiology* **278(6)**: 2039–2049.

Roh SC, Park EJ, Shim M, Lee SH. 2016. EEG beta and low gamma power correlates with inattention in patients with major depressive disorder. *Journal of Affective Disorders* **204**: 124–130. DOI: 10.1016/j.jad.2016.06.033.

Shahaf G, Yariv S, Bloch B, Nitzan U, Segev A, Reshef A, Bloch Y. 2017. A Pilot Study of Possible Easy-to-Use Electrophysiological index for Early Detection of Antidepressant Treatment Non-response. *Frontiers in Psychiatry*, **8**: Article 128. DOI: 10.3389/fpsyt.2017.00128.

Spasić S, Kalauzi A, Culić M, Grbić G, Martać Lj. 2005. Fractal analysis of rat brain activity after injury. *Medical and Biological Engineering and Computing* **43**: 345–348. DOI: 10.1007/BF02345811.

Stam CJ. 2005. Nonlinear dynamical analysis of EEG and MEG: a review of emerging field. *Clinical Neurophysiology* **116(10)**: 2266–3211.

Stewart JL, Bismark AW, Towers DN, Coan JA, Allen JJB. 2010. Resting frontal EEG asymmetry as an endophenotype for depression risk: sex-specific patterns of frontal brain asymmetry. *Journal of Abnormal Psychology* **119:** 502–512. DOI: 10.1037/a0019196.

Stokić M, Milovanović D, Ljubisavljević M, Nenadović V, Čukić M. 2015. Memory load effect in auditory-verbal short-term memory task: EEG fractal and spectral analysis. *Experimental Brain Research,* **233(10)**: 3023–3038. DOI: 10.1007/s00221-015-4372-z.

Vapnik VN. 1988. *Statistical Learning Theory* (chapter 10, p. 42). John Willey and Sons Inc., Canada.

Van der Vinne N, Vollebregt MA, van Putten MJAM, Arns M. 2017. Frontal alpha asymmetry as a diagnostic marker in depression: fact or fiction? A meta-analysis. *Neuroimage: Clinical* **16**: 79-87.

Vederine FE, Wessa M, Leboyer M, Houenou J.2011. A meta-analysis of whole-brain diffusion tensor imaging studies in bipolar disorder. *Progress in Neuro-Psychopharmacology & Biological Psychiatry* **35**: 1820–1826. DOI: 10.1016/j.pnpbp.2011.05.009.

Unnikrishnan P, Kumar DK, Arjunan SP, Kumar H, Mitchell P, Kawasaki R. 2016. Development of Health Parameter Model for Risk Prediction of CVD Using SVM, *Computational and Mathematical Methods in Medicine*, 3016245. DOI 10.1155/2016/3016245

Werner NS, Meindl T, Materne J, EngelRR, Huber D, Riedel M, Reiser M, Henning-Fast K. 2009. Functional MRI study of memory-related brain regions in patients with depressive disorder. *Journal of Affective Disorders* **119(1–3)**: 124–131. DOI: 10.1016/j.jad.2009.03.003.

Witten IH, Frank E. 2005. *Data Mining: Practical Learning Tools and Techniques* (2$^{nd}$edition), Elsevier, pp. 90-97.

World Health Organization. 2017. Depression and other common mental disorders. http://apps.who.int/iris/bitstream/10665/254610/1/WHO-MSD-MER-2017.2-eng.pdf

World Health Organization. 1993. ICD-10 Classification of mental and behavioral disorders. Clinical descriptions and diagnostic guidelines.

World organization of Mental Health. 2012. Depression: a Global Crisis; World Mental Health Day, October 10, 2012 http://www.who.int/mental_health/management/depression/wfmh_paper_depression_wmhd_2012.pdf

**Acknowledgment:** We would like to thank Christopher Stewart, PhD, analytical linguist from Google for help in improving the text and Dr. Tomasz Smolinski from Delaware State University for useful discussions.